# A Decision-Analytic Model for Using Scientific Data

Harold P. Lehmann
Section on Medical Informatics, MSOB x215
Stanford University, Stanford, CA 94305–5479
(415) 723–2954, lehmann@sumex-aim

Many Artificial Intelligence systems depend on the agent's updating its beliefs about the world on the basis of experience. Experiments constitute one type of experience, so scientific methodology offers a natural environment for examining the issues attendant to using this class of evidence. This paper presents a framework which structures the process of using scientific data from research reports for the purpose of making decisions, using decision analysis as the basis for the structure, and using medical research as the general scientific domain. The structure extends the basic influence diagram for updating belief in an object domain parameter of interest by expanding the parameter into four parts: those of the patient, the population, the study sample, and the effective study sample. The structure uses biases to perform the transformation of one parameter into another, so that, for instance, selection biases, in concert with the population parameter, yield the study sample parameter. The influence diagram structure provides decision theoretic justification for practices of good clinical research, such as randomized assignment and blindfolding of care providers. The model covers most research designs used in medicine: case-control studies, cohort studies, and controlled clinical trials, and provides an architecture to separate clearly between statistical knowledge and domain knowledge. The proposed general model can be the basis for clinical epidemiological advisory systems, when coupled with heuristic pruning of irrelevant biases; of statistical workstations, when the computational machinery for calculation of posterior distributions is added; and of metaanalytic reviews, when multiple studies may impact on a single population parameter.

## 1 The Problem

Decision-analytic models have been applied to different areas in artificial intelligence, including diagnosis (Heckerman, Horvitz, & Nathwani, 1989), learning (Star, 1987; Buntine, 1987), vision (Levitt, et al., 1988), and control of inference (Horvitz, 1989; Breese & Fehling, 1988). An activity common to these models is the agent's updating his belief in relevant propositions on the basis of evidence. The models usually leave implicit the decision maker's belief in the method by which the data were obtained. To be done properly, updating must include this belief in each of many possible models of observation of the data. We shall show that, in certain contexts, the space of all possible observational contexts can be parameterized to be both assessable and computable. Scientific research is one such context; specific biases can be used to provide the necessary parameterization.

Parameterization of the observational contexts is domain-dependent at different levels of the meaning of *domain*. We use the following hierarchy. The topmost level is the general field of scientific or systematic observation. The second level is the field of research aimed at discerning causal relations, as opposed, say, to exploratory descriptions. A third level is that of medical research, which eliminates from consideration a number of destructive experimental designs. A fourth level is the class of study, including, for instance, balanced-design, case-control, cohort, or randomized, studies. A fifth level is the object (domain of interest), such as cardiology. Our goal is to offer a structure that allows parameterization of all studies within the fourth domain. A successful parameterization at this level should enable us to handle a wide variety of medical (object) domains, while the structure will probably be effective at higher levels as well.

We may visualize the problem using the influence diagrams (Howard & Matheson, 1981) of Figure 1. Figure 1a shows that the decision maker makes his decision, $D$, knowing the data, $x$, at the time of the decision, and thinking about the effect of the parameter, $\theta$, on relevant outcomes, $\Omega$, which affect the value, $V$, of the decision maker. The calculation of expected utility for the decision requires the prior probability distribution for the parameter $\theta$, in addition to the preposterior distribution $P(x \mid \theta)$. This formulation, however, leaves out the possibility that the data were obtained in different ways, which is equivalent to its preventing the decision maker from modeling his uncertainty in the preposterior distribution. In Figure 1b, we include the observational context explicitly, operationalized (or parameterized) in terms of a bias parameter, $\phi$, and make explicit the types of data, $y$, that bear on $\phi$, separately from those that bear on $\theta$. To use this structure, we need the prior over $\phi$ and the preposterior distribution, $P(x \mid \theta, \phi)$. The latter distribution contains our knowledge about the mechanism by which the data are obtained in a particular experimental context.

As an example of the importance of the observational context, consider the following problem. I have two coins, one of which I may bring to a gambling event. I would, of course, prefer the coin that gives me the greatest odds of winning me the most money. The coins are apparently identical, but they may have different chances of landing heads. I flip each coin 100 times, asking an assistant to give me the coins alternatingly. This process gives me two lists of tosses and outcomes, one for each coin. Most utility functions would have me choose the coin with the higher number of heads as the coin to take (unless that proportion is *too* high, in which case I risk being found

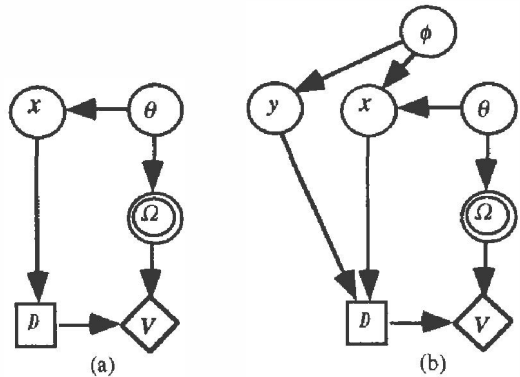

- $\theta$    Parameter of interest
- $x$    Data pertaining to parameter of interest
- $D$    Decision
- $\Omega$    Outcomes
- $V$    Value
- $y$    Data pertaining to bias parameter
- $\phi$    Bias parameter

Figure 1. General influence-diagram models for belief updating. (a) Using primary data to update belief in the parameter of interest. (b) Parameterizing the observational context and using additional data to update belief in the observational model.

out as a cheater), based on the binomial likelihood: for $i = 1,2$, the posterior probability of $\theta_i$ (the chance of coin $i$ falling heads) given $x_i$ (the observed number of heads from flipping coin $i$) is proportional to $\theta_i^{x_i}(1-\theta_i)^{100-x_i}$, and this might be your advice to me. But what if I told you that my assistant is three years old? You might reconsider your advice, since the assumption that each list pertains to a separate coin is now probably invalid. We should introduce two new parameters, $\phi_i$, which represent the proportion of the time our young assistant gave me coin $i$, but told me it was the other one. The correct model for the preposterior function for $x_1$ now is

$$l(x_i \mid \theta_i') = \theta_i'^{x_i}(1-\theta_i')^{100-x_i},$$

where $\theta_i'$, the *effective* parameter, is a function of $\theta_1$, $\theta_2$, $\phi_1$, and $\phi_2$. Specifically,

$$\theta_i' = \theta_1(1-\phi_1) + \theta_2\phi_2.$$

To find the posterior distribution of $\theta_i$ given the data, we must integrate out the observational parameters. Alternatively, we could ask, what shape of a prior over the observational parameters would lead to a change in action?

## 2 A Model for the Use of Reported Scientific Data

We narrow our attention to study designs for medical research, at the fourth level of our domain hierarchy, for a number of reasons. First, providing data for high-stakes decisions relating to individual situations is a major purpose of medical research, as evidenced by the large number of clinical studies. Second, while our discussion will remain general, it will be focused by our considering a narrower domain than all scientific research, thereby clarifying the model. Third, the concept of bias in medical research is well established, giving us a baseline against which to judge the external validity of our model.

### 2.1 Details of the General Model

Our general model, depicted in Figure 2, is an expansion of Figure 1, based on Shachter, et al. (1989). It consists of a central framework with peripheral adjustments. The framework allows the representation of a number of modeling decisions, separate from the domain issue of which therapy to implement. We will point out these modeling decisions as they come up. One general approach for making modeling decisions is for the physician to explore each potential model resulting from one modeling decision, and integrate expected value across these models, weighted by a prior probability on each model. A second approach is for the physician to perform a sensitivity analysis across some or all models, and either continue with a single best model, or integrate across models, depending on the outcome of the analysis.

We now circumnavigate our general model, Figure 2. We have made explicit that the physician is concerned about a particular *Outcome of Interest*, such as mortality, morbidity, or cost. The fact that he uses certain parameters to help him assess the likelihood of that outcome occurring is a result of a separate decision as to the probabilistic model he chooses to use.

We have split the general *Parameter of Interest*, $\theta$, from Figure 1 into four separate parameters: the *Patient Parameter*, $\theta_{pt}$, the *Population Parameter*, $\theta_{pop}$, the *Sample Parameter*, $\theta_{sample}$, and the *Effective Sample Parameter*, $\theta'_{sample}$. By separating $\theta_{pt}$ from $\theta_{pop}$, we allow the physician to take two types of modeling actions. One is for the physician to use the same data for many different clinical situations, if he knows how to derive the distribution for $\theta_{pt}$ from *Hyperparameters* updated by $\theta_{pop}$ in a hierarchical Bayes framework (Berger, 1985), which may require traditional statistical models. A second type of modeling action is for him to effect a *metaanalysis* (L'Abbé, 1987), using multiple sources of data (several different papers) to update his

209

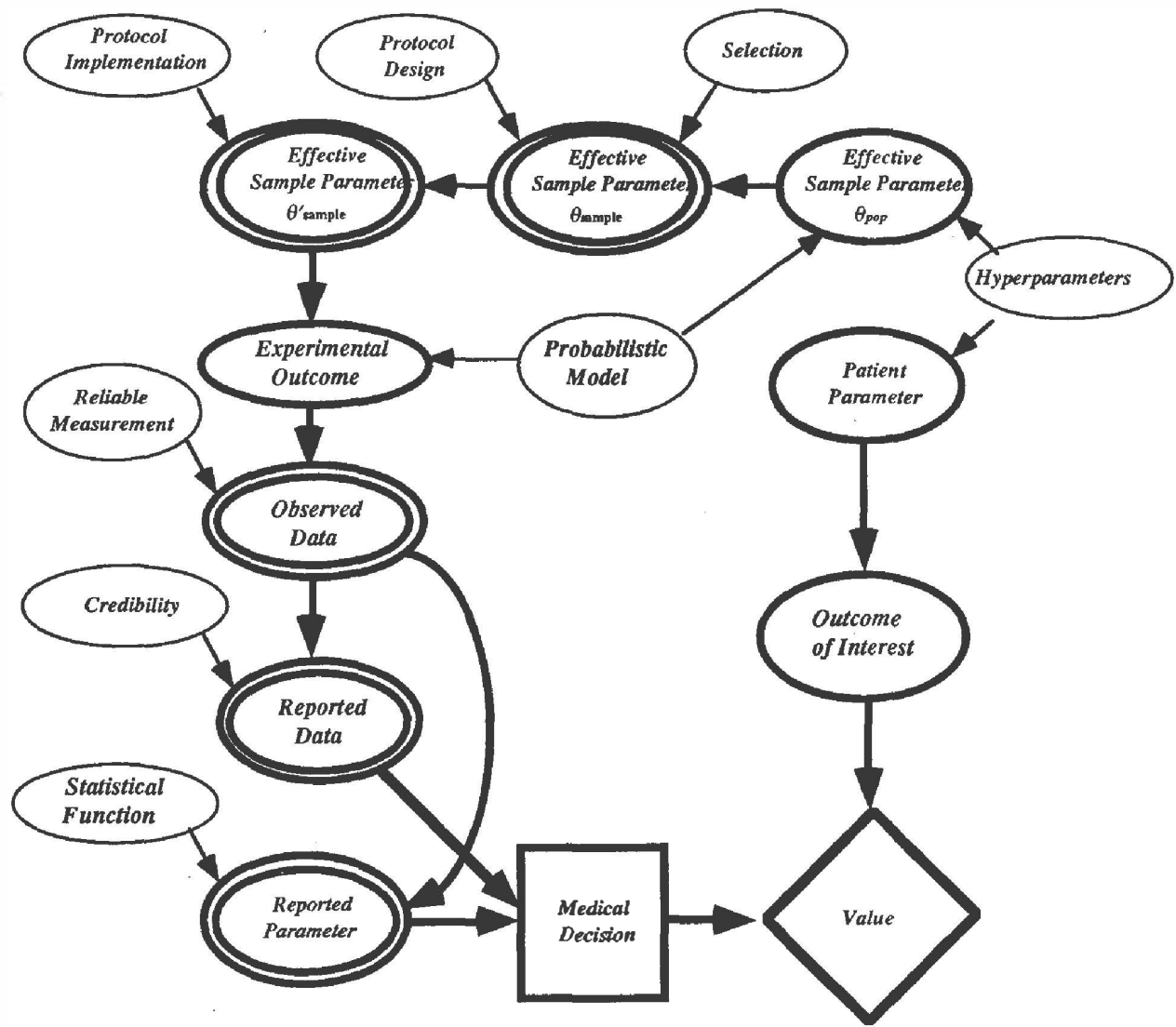

Figure 2. Influence-diagram model for using research data in the general medical decision-making problem.

belief in $\theta_{pop}$. By separating $\theta_{sample}$ from $\theta_{pop}$, we allow for the physician to generalize the study from $\theta_{sample}$, the belief in the parameter for the perfectly-done study, to $\theta_{pop}$, the belief in the parameter for the ideal population. The physician could calculate $\theta_{sample}$ exactly from $\theta_{pop}$, if he knew how the *Selection* from the general population led to the patient pool sampled in the study and by what *Protocol Design* the patients were assigned their respective interventions (which is why $\theta_{sample}$ is a deterministic node). The last parameter, $\theta'_{sample}$, contains the physician's belief in the parameter of the study as performed. If the physician knew exactly how $\theta_{sample}$ was modified by the experimental *Protocol Implementation*, he could calculate exactly what the parameter was to which the experimental outcome refers.

The *Data* node of Figure 1 has been split into an *Experimental Outcome* node, representing the actual results from the study as performed; an *Observed Data* node, representing the data recorded by the experimenters; and a *Reported Data* node, representing the data as reported in the written article and the only data available to the reader. The likelihood of the experimental outcome given the *Effective Sample Parameter* is the probability distribution that depends on the *Probabilistic Model* the physician chooses for modeling the outcome of the experiment. For mortality studies, as an example, this model is usually binomial.

If the physician knew the *Measurement Reliability* of the recording instruments, say the true sensitivity and specificity, and the true model relating the reliability to the actual and the observed data, he could calculate exactly the *Observed Data* from the *Experimental Outcome*. Similarly, if he knew exactly the *Credibility* of the reporters, he could calculate the *Reported Data* from the *Observed Data*.

Researchers often report in their papers only their estimate of the parameter, rather than all the observed data. This estimate is a *Statistical Function* of the observed data, such as a mean or a regression coefficient. Statistical decision theory (Berger, 1985) concerns itself with the choice of that function. We note that, without at least the observed data, we lose information that could potentially alter our decision. The sufficient statistics proposed by statistical decision theory, then, appear to be context dependent, and our model makes explicit just this context.

### 2.2 An Example of Navigating the Model

To illustrate how all this might work in a specific instance, we consider a 55 year old white woman who has just had a heart attack and who has been brought into the hospital almost immediately after symptoms of chest pain, nausea, and sweating set in. Her physician, besides needing to stabilize her acute cardiovascular status, wants to prevent worsening of her general cardiac condition. The doctor knows that a drug, metoprolol, is considered possibly able to do so. He is primarily concerned with minimizing the patient's chance of death, and thereby maximizing the heart attack victim's life expectancy. There are some known side effects of the medication. Should he use the drug?

Let us go through an abbreviated analysis using a study (Hjalmarson, et al., 1980) that bears on this question. As discussed before, we leave the choice of the value model to be implicit. The *Outcome of Interest* is mortality. We make the modeling decision to use the patient's probability of death as the *Patient Parameter*, which we will assume to be constant over time (constant hazard model). For the modeling decision of the choice of referent *Population*, we have at least two choices on the basis of cardiology-domain knowledge: middle-aged women and middle-aged adults. If we choose the population of both sexes, there will be a larger number of studies, each with a large sample size, that we can bring to bear on this problem, which, in concert, may affect the *Hyperparameters* as much as if we used only the subgroup of women only. Clearly, there is a modeling decision trade-off between specificity of the data versus its the amount of data available. For the purposes of this paper, we shall take the modeling action of using only the combined population of middle-aged adults.

The *Sample* in the study consists of all heart-attacks victims from south Sweden in the late 1970s. This characterization represents *Selection* from our population on ethnic grounds, but not on the basis of referral, diagnostic purity, or diagnostic access biases (Sackett, 1979). The *Protocol Design* is reported to have been that of a double-blinded and randomized clinical trial, in which the assignment of a patient to a given treatment is independent of the patient's baseline status. There is evidence to support the claim that the *Protocol Implementation* was identical to the design. For instance, the compositions of the metoprolol and placebo groups are similar with respect to relevant characteristics, based on reported baseline data, corroborating the implementation of randomization. The number of withdrawals from the study on the basis of side-effects is also similar between the two groups, suggesting that if there were some unblindfolding of the care providers such that the treating physicians became aware of the true treatment assignments, its degree was the same in both groups.

Estimating the actual degree of withdrawal explicitly is important for calculating a posterior distribution on $\theta_{sample}$. Withdrawal refers to a patient's not receiving the treatment to which he was assigned. This estimation adds a *bias parameter* to be inferred, which, in turn, results in our considering a space of observational models larger than we would be considering without including the withdrawal bias. The withdrawal bias parameter in this study models the fact that the effective sample parameter for metoprolol was a result of mixing the treatment group with a third group of patients receiving no treatment, that is, a group with the baseline mortality risk (the group of patients withdrawn). The new parameter to be inferred is the degree of mixing, $\phi$. Shachter and colleagues (1989) offer a mathematical form for the effective sample parameter,

$$\theta'_{sample, metoprolol} = (1 - \phi) \cdot \theta_{sample, metoprolol} + \phi \cdot \theta_{sample, baseline},$$

showing that $\theta'_{sample}$, is a function of $\theta_{sample}$ and of $\phi$, as represented by the deterministic node in Figure 2. We note the similarity to our coin problem, in Section 1, where now, $\phi_1 = \phi_2 = \phi$. In the study, the reported overall withdrawal rate is 19.1 per cent in both groups, and we could use this to update our belief in $\phi$.

Continuing around Figure 2, we find that the investigators use the binomial *Probabilistic Model*, in keeping with our definition of the parameter of interest. *Measurement Reliability* depends on the sensitivity and the specificity of the sensing mechanisms. For mortality studies, the sensitivity, P*(labeling patient as "dead" | patient deceased)*, and specificity, P*(labeling the patient as "alive" | patient is alive)*, depend on patients who have dropped out of the study. The authors assure us that the mortality status of each patient entered into the study was assessed, regardless of subsequent treatment status, and their credentials are such that we consider them to have high *Credibility*. Finally, the authors report both mortality rates and life tables as their *Statistical Functions* of the data.

## 3 Using Bias to Parameterize the Space of Clinical Studies

We now have the basic architecture for parameterizing the space of observational models through expanding the peripheral nodes of Figure 2. We gave the details of one instance in the mixture model resulting from patients withdrawn from the study. Our present concern is simply to locate all potential biases in a single model. We wish, therefore, for as comprehensive as possible a set of biases that spans the space of all studies. To show this set, we need to examine more closely the top line of Figure 2: the relationships among $\theta_{pop}$, $\theta_{sample}$, and $\theta'_{sample}$. In causal studies, each of these parameters is a probability (e.g., mortality risk), or a parameter of a probability distribution, and hence represents an abstraction of a local node group of an influence diagram. The most general form of such a local node group is given in Figure 3: the node *Patient's Event* is dependent on the nodes *Patient's Baseline State* and *Patient's Exposure to an Agent of Interest*. For instance, $\theta_{sample}$ is defined by the probability distribution of the events observed in the sample patients, given the baseline states of the sample patients and the agents to which those patients were exposed. Similarly for the population and for the effective sample. The language comes from Feinstein (1985), who uses this local node group to account for the widest possible range of initial states (e.g., healthy, diseased), agents (e.g., environmental exposure, medical therapy, process intervention), and events (e.g., mortality, pain, contraction of disease), thereby allowing for the same structure to be used in analyzing the entire class of comparative studies. Thus, to explore the relationship between specific biases and the parameter in which we are interested, we need to examine the primary belief network of which the parameter is an abstraction, and to locate the dependencies of the parameter on those biases. Then, the relationship between two parameters can be worked out at the level of these primary local node groups.

In Figure 4, we present a fragment of such an expanded belief network, showing the interplay between different classes of bias and the relationship between the *Sample Parameter* and the *Effective Sample Parameter*. The citation sources for the biases are indicated. As an example, we note that *Classification Error* is a class of bias upon which the assessment of *Effective Sample Initial State* is dependent. There is a large list of specific biases which fall into this category, such as *Previous Opinion Bias* and *Diagnostic Suspicion Bias*. These biases lead to an difference between the *Sample Initial State* and the *Effective Sample Initial State*; this difference leads, in turn to a difference between the *Sample Parameter* and the *Effective Sample Parameter*. This fragment, then, is the mechanistic interpretation (or, expansion) of the arc in Figure 2 between the *Sample Parameter* and the *Effective Sample Parameter*.

We note that the model incorporates a wide variety of methodological elements, such as misclassification, misassignment, and conduction of the study (blindfolding) all into a single model. This analysis of mechanism allows us to preserve (or define) the semantics of the biases, while providing a numerical environment in which to use them.

## 4 Previous Attempts to Model the Use of Scientific Data

Rennels (1986) uses AI heuristics in constructing a system, ROUNDSMAN, that offers therapeutic suggestions for a particular patient, on the basis of articles in its knowledge base. The primary heuristic is the calculation of the "distance" of a paper to a particular domain decision. For the program to function, a domain expert must pre-process each article; the user enters the defining characteristics for the patient. Problematic in Rennels' approach is that all types of biases and adjustments are combined into a single score, without an underlying structure like the one we have developed. Furthermore, by not considering the probabilistic model for the data, ROUNDSMAN is unable to give an integrated numeric solution to the question of what the physician should do on the basis of the data.

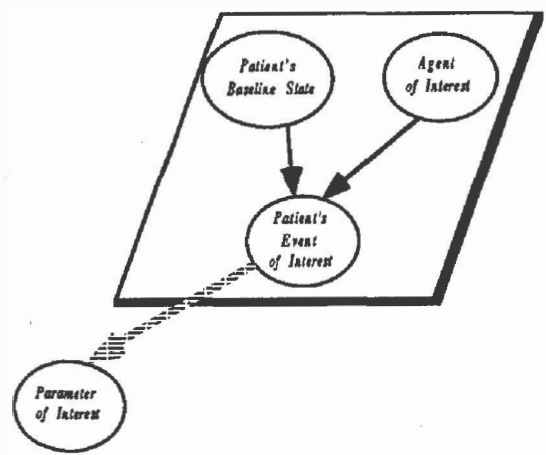

Figure 3. The general local node group of which a parameter of interest (such as the *Population Parameter*) represents the indicated probability distribution After Feinstein (1985).

Our model is in the same spirit as Eddy's confidence profile method (Eddy, 1989). His approach is motivated more by the desire to perform metaanalyses than the need to analyze in depth a single article. His "chains" are concerned with the causal connections between parameters and outcomes of interest. He deals with biases as correction factors, similar to our peripheral nodes in Figures 2.

The goal of the REFEREE research endeavor (Lehmann, 1988) is to create a system that aids a user in evaluating the methodological quality of a paper. The REFEREE team first implemented that goal in a rule-based program. Because of semantic and representational deficiencies, they redefined the goal to be the replicability of the conclusion of the study, and implemented the new goal in a probabilistic program and in a multiattribute-value-based program. However, even with these normative bases, it is not clear what should be done with the output of these systems. The approach we have taken in this paper is an answer to that question.

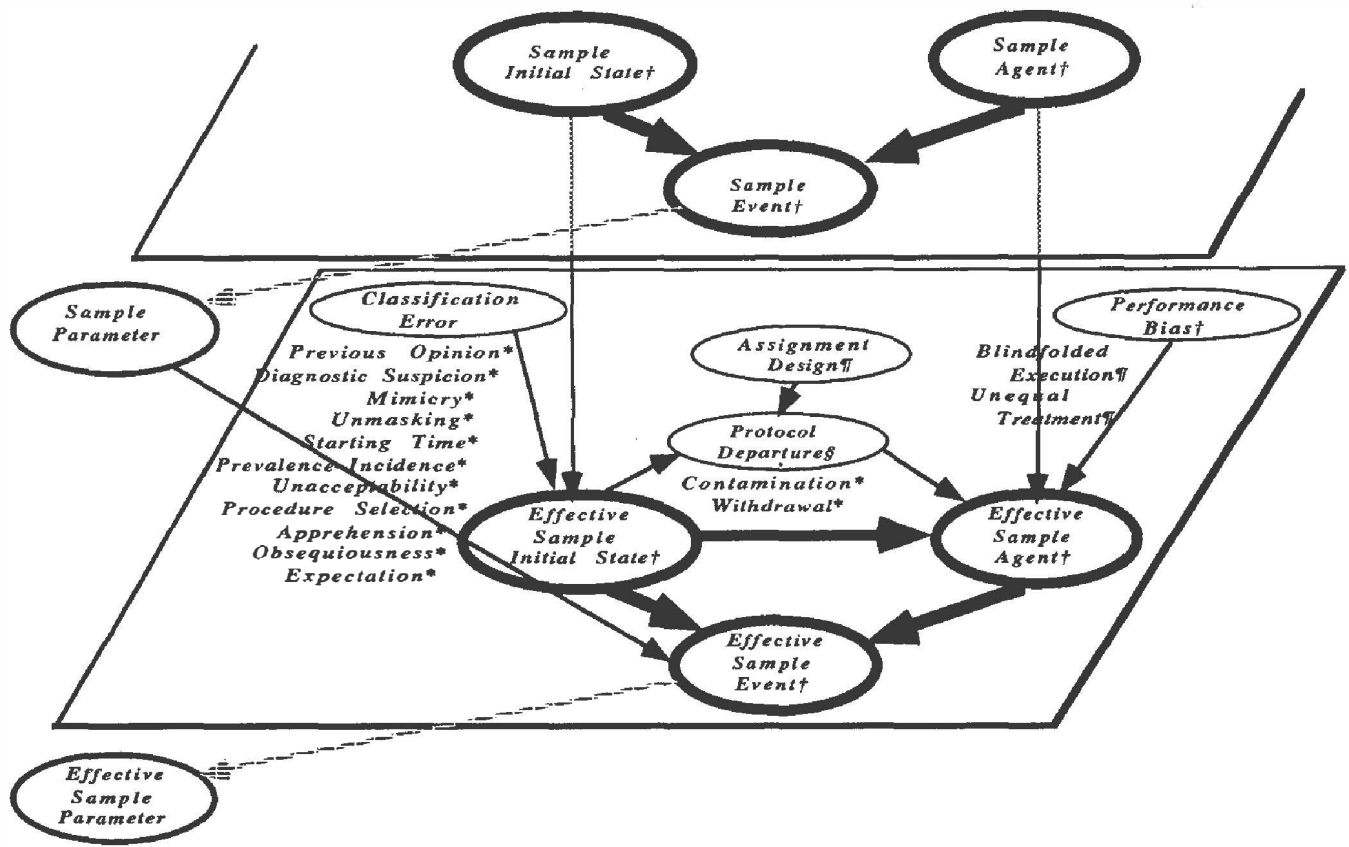

Figure 4. Primary influence diagram, defining the *Effective Sample Parameter*, and its relationship to the *Sample Parameter*. Thickly drawn nodes and arcs indicate the local node groups of Figure 3. The horizontally hatched arcs indicate that the destination node is an abstraction of the probability distribution of the destination node. Footnote symbols indicate citation source for biases: * Sackett (1979); † Feinstein (1985); ¶ Lehmann (1988).

Finally, the field of clinical epidemiology is "concerned with studying groups of people to achieve the background evidence needed for clinical decisions in patient care" (Feinstein 1985, p. 1). Besides bringing epidemiological and statistical techniques to the medical bedside, clinical epidemiologists are concerned with physicians' coherent reading of the clinical research literature. Sackett, Haynes, and Tugwell (1985) have published a *sequential* algorithm for using the literature: If the article fails any step along the path, it is eliminated from consideration. L'Abbé, Detsky, and O'Rourke (1987) describe sequential steps to be taken for using an article in a metaanalysis. Their measure of quality is based on the rating-sheet approach of Chalmers and colleagues (1981). Our approach, in contrast, is *global* in that an article is implicitly discarded only after the probability distribution, $P(\theta_{pt} \mid Reported\ Data)$, is determined; a distribution close to the prior distribution over $\theta_{pt}$ suggests that the biases and adjustments have washed out any meaningfulness from the reported data. The primary advantage of the global approach is that it does not explicitly throw out data, unlike the sequential algorithm. Although a study may be designed and executed poorly or may be somewhat irrelevant, its data may still have an important bearing

213

on the calculation of expected value. The disadvantage of the global approach is its computational complexity.

## 5 Uses of the Model

The model may be used in several ways. At the least, it may help a reader to organize the study of research results. This use leads to a qualitative analysis that, alone, may result in all the insights the user may want. We focus our comments on computer applications, however. A quantitative approach helps a user to organize the clinical epidemiological concerns and provides the basis for a decision support system and statistical workstation, where the resulting probability distributions may be displayed and studied, and where distributions over probability models may be considered as well.

In using the model to construct decision-support systems, we must construct models of how evidence in the study report updates our belief in a model. Extending the model results in our building a large knowledge base of research designs. With our general layout of the space of studies, it becomes easier to accrue new biases and research designs. The model we have given serves as the core, the methodological and statistical knowledge base (Figure 5). Domain knowledge is incorporated as new conditions on the biases (new predecessors), or new sources of evidence for the different biases (new descendents). In using the model, we begin with the knowledge base of biases, and use either methodologic or object-domain knowledge heuristically to prune that KB on the basis of which biases are irrelevant for the context of interest. With this reduced model, we then expand the possible evidential nodes for the bias parameters and search for evidence in the study report that bears on the bias parameters. Once we have updated our belief in those parameters, we examine the primary data themselves, which update both the bias parameters and the domain parameter. Finally, we remove the bias parameters as nuisance parameters. Shachter (1988) gives the general approach for this, and Barlow, Irony, and Shor (1988) give an example pertaining to experimental designs. This removal leaves us with a posterior probability distribution on the *Population Parameter*. We then derive the distribution for the *Patient Parameter* and calculate the expected utility for each action. We may do the reverse as well, and ask: what *prior* probability distribution would have us take different action?

We have presented an architecture for a large knowledge base of science methodology. The details of this knowledge base have implications for designing scientifically (rationally) functioning computer agents which collect data about the real world. The model we have presented eliminates most theoretical problems with previous approaches and provides a framework for future work. Extensions of our ideas include generalizing to other evidential domains in the hierarchy of domains presented at the beginning of this paper.

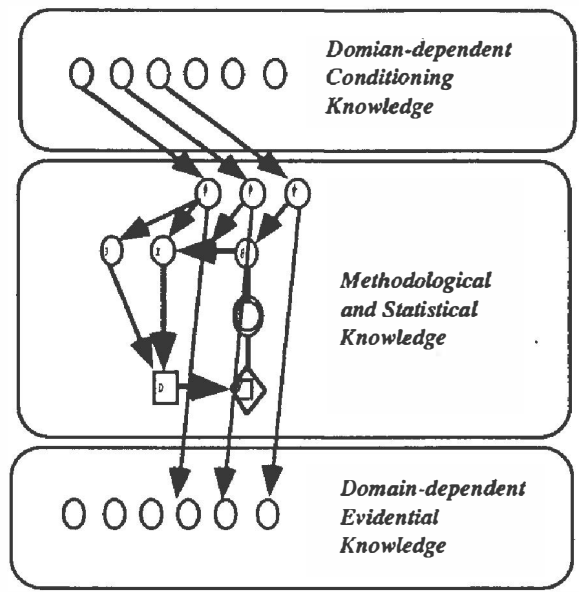

Figure 5. General architecture for a decision support system. The central knowledge base contains methodological and statistical knowledge of biases and of the decision process (as in Figures 1, 2, and 3). The upper knoweldge base contains domain-specific knowledge regarding conditioning knowledge for specific biases, while the lower one contains domain-specific knowledge about what sorts of information from a study are evidential for specific biases.

We also need to make more explicit the various modeling decisions the reader undertakes in performing the type of analysis we have been discussing. The work we present here can be followed now by a practical instantiation.

### Acknowledgments

Thanks to Greg Cooper and to Ross D. Shachter for help in formulating these ideas. Sources of funding include Grant LM-04136 from the National Library of medicine, and NIH grant RR-00785 for the SUMEX-AIM computing facilities.### References